# OAPA : Une Ontologie pour la description de l'Activité des Personnes Agées

Rahma DANDAN* - Sylvie DESPRES* - Jérôme NOBECOURT*

*Université Paris 13, LIMICS, UMR_S 1142, INSERM*
*74 rue Marcel Cachin*
*93017 Bobigny cedex*
`*{rahma.dandan,sylvie.despres,jerome.nobecourt}@univ-paris13.fr`

RÉSUMÉ. *L'amélioration de la pratique d'activités et le changement de mode de vie sédentaire constitue une priorité de santé publique et fait partie des stratégies mondiales de santé (OMS). Dans ce contexte, la mise en place d'un système de suggestions d'activités adaptées au profil gérontologique des personnes âgées (PA) suscite un intérêt majeur. Dans ce papier, nous présentons l'approche utilisée pour modéliser l'activité et concevoir l'Ontologie de l'Activité pour les Personnes Agées (OAPA). Parmi les approches utilisées pour modéliser les connaissances, l'approche axée sur les connaissances a été utilisée dans ce travail. Cette approche est la plus efficace pour représenter l'activité, gérer les données de capteurs et raisonner sur les connaissances modélisées. En utilisant des méthodes d'ingénierie des connaissances, nous avons pu modéliser l'activité en considérant les variables entrant dans la définition du profil des PA. Pour concevoir OAPA, nous nous sommes fondés sur des scénarios et des questions de compétences comme étant un moyen pour déterminer les spécifications de l'ontologie à concevoir et d'évaluer notre modèle de connaissances. L'approche ontologique utilisée permettra de raisonner en utilisant l'ontologie OAPA pour établir des suggestions personnalisées, ciblées et incitatives pour les PA sédentaires, dépendantes, en perte d'autonomie ou nécessitant une telle suggestion.*

ABSTRACT. *Improving the practice of activities and changing sedentary lifestyle is part of global health strategies (WHO). In this context, the establishment of an activity suggestion system arouses a major interest. In this paper, we present the approach used to model the activity and to design of Ontology of Activity For the Elderly (OAFE). Among the approaches used to model the activity, the knowledge-based approach has been used in this work as being the most effective for representing activity, managing sensor data, and reasoning about modeled knowledge. Using knowledge engineering methods, we were able to model the activity concept by considering the variables used in the definition of the elderly profile. To design OAFE, we relied on scenarios and competency questions as a means to determine ontology specifications to design and evaluate our knowledge model. The ontological approach used will make it possible to reason using the OAFE ontology to establish personalized, targeted and incentive suggestions for elderly.*

MOTS-CLÉS : *Ontologie, Modélisation, Conceptualisation, Activité physique, Personnes âgées.*

# 1. Introduction

Face à une population vieillissante, les enjeux de santé deviennent de plus en plus cruciaux. Ils consistent essentiellement à tenter de prévenir de nombreuses pathologies et conjointement à améliorer la qualité de vie des personnes âgées (PA). En effet, le vieillissement modifie les capacités à s'alimenter, altère l'état de santé physique et mental, augmente le niveau de dépendance et diminue l'environnement social et économique (Escalon *et al.*, 2010). Tous ces changements participent entre autre à l'augmentation du risque d'inactivité, de sédentarité et à la prévention des facteurs de risque qui y sont liés.

Quelle que soit l'activité exercée, elle participe à la prévention primaire et secondaire des maladies chroniques. Ainsi, l'amélioration de la pratique d'activités et le changement de mode de vie sédentaire constitue une priorité de santé publique et fait partie des stratégies mondiales de santé (OMS). Cependant, leurs stratégies, principalement présentées sous forme de fascicules de recommandation d'activités physiques, sont loin d'être incitantes et adaptées au profil et à l'état de santé des PA.

Pour participer aux changements des pratiques physiques et aider les PA à vivre de manière indépendante, de nombreuses stratégies ont émergées ces dix dernières années visant soit à identifier les activités des PA, soit à acquérir leurs préférences afin d'adapter la suggestion d'activité à leurs envies. Dans ce contexte, il est utile de mettre en place un système de suggestions d'activité adaptées au profil gérontologique des PA.

L'approche ontologique représente une solution efficace pour favoriser le partage, l'intégration et la gestion de l'information de ressources hétérogènes. Cette approche représente également un moyen efficace pour définir un vocabulaire commun au domaine d'étude et pour spécifier la sémantique des activités et disposer d'une base de connaissances flexible, extensible et interopérable.

Ce travail s'inscrit dans un projet visant à développer un système automatisé de suggestions d'activités selon le profil de la PA. Il est indispensable de définir le profil de la PA correspondant au mieux à son état de santé, ses envies, ses objectifs, ses motivations, ses réticences et à l'environnement de la PA pour adapter la suggestion et permettre une prise en charge personnalisée. Le recueil de données constituant le profil d'activité de la PA est réalisé au moyen d'objets connectés. L'objectif de ce travail est de fournir un cadre méthodologique pour la modélisation et la conception d'une ontologie de l'activité.

Le papier est organisé de la manière suivante : (i) une étude comparative des différents travaux et des différentes approches utilisées afin de motiver notre volonté de modéliser et concevoir une ontologie de l'activité au sein de notre projet ; (ii) une modélisation conceptuelle de l'activité et de la personne dans un contexte de suggestion d'activité ; (iii) une présentation de l'ontologie OAPA ; (iv) une évaluation de l'ontologie OAPA ; (v) une conclusion succincte du travail présenté.

## 2. Etat de l'art

La représentation des connaissances d'un domaine particulier à partir de modèles conceptuels constitue une étape importante dans le processus de développement de système traitant un volume important de données (Batani *et al.*, 1992). La modélisation relative à l'identification d'activités a bénéficier ces dix dernières années de recherches approfondies s'appuyant sur deux types d'approches, une approche axée sur les données et une approche axée sur les connaissances. La première approche permet d'analyser et de modéliser des données de capteurs sur des périodes à long terme et de traiter des données à grande échelle. Elle nécessite la génération d'un modèle pour chaque activité en utilisant des stratégies d'apprentissage probabilistes et statistiques (Gayathri *et al.*, 2017). La seconde approche, axée sur les connaissances, utilise des méthodes d'ingénierie des connaissances pour modéliser les activités en utilisant des connaissances de domaine et des techniques de raisonnement fondées sur l'intelligence artificielle pour être en mesure d'établir des inférences (Yu *et al.*, 2015). Cette approche est la plus utilisée et la plus efficace pour représenter l'activité, gérer les données de capteurs considérés comme des ressources hétérogènes et raisonner sur les connaissances modélisées. En effet, les ontologies sont de plus en plus utilisées au sein de systèmes complexes pour modéliser et raisonner sur les activités, quelles soient simples ou composites. Certains auteurs (Chen *et al.*, 2008 ; Riboni and Bettini, 2011 ; Wongpatikaseree *et al.*, 2013) utilisent les ontologies pour identifier les activités de la vie quotidienne (par exemple, faire le ménage, préparer un repas, etc.) dans le contexte de *Smart Home* qui dispose de capteurs (*sensors*) détectant des tâches contextuelles (remplir la casserole d'eau, allumer la plaque de cuisson, mettre la casserole au feu, etc.) *via* les définitions contenues dans l'ontologie. D'autres auteurs (Wieringa *et al.*, 2011 ; Pramono *et al.*, 2013 ; Nassabi *et al.*, 2014 ; Zolfaghari *et al.*, 2016) utilisent les ontologies pour suggérer des activités ou pour la reconnaissance et la suggestion d'activités simultanément (Filos *et al.*, 2016 ; Ni *et al.*, 2016). Toutefois, la plupart échouent dans la personnalisation de la suggestion car tous les paramètres influençant la suggestion ne sont pas pris en considération.

L'approche prédominante dans ces études consiste à dissocier l'ontologie noyau (*core ontology*) fournissant les concepts structurant du domaine et décrivant les relations entre ces concepts, de l'ontologie de domaine fournissant les concepts spécifiques du domaine d'étude ainsi que les relations entre ces concepts. Néanmoins, la réutilisation de ces ressources est complexe. En effet, aucune ontologie des auteurs précédemment cités n'est disponible pour être réutilisée. Parmi les ontologies incluant le concept d'activité « *Activity* », certains auteurs mettent à disposition leur hiérarchie de classes sans donner les liens entre ces classes. Quelques unes de ces ontologies légères et méta-thesaurus sont représentés dans la Figure 1. D'autres auteurs fournissent leur modélisation de l'activité : - ADL Ontology (Chen *et al.*, 2008) ; - Palspot Project Ontology (Riboni and Bettini, 2011) ; - Activity Log Ontology (Wongpatikaseree *et al.*, 2013) ; - modèle de conception d'ontologie générique (Abdalla *et al.*, 2016) ; - ECOPPA (Hoda *et al.*,

2018). Dans la plupart des cas, ces représentations sont générales, incomplètes ou encore dépendantes d'un contexte.

Le Tableau 1 résume les principaux termes de premier niveau reliés au concept « *Activity* » qui ont été identifiés dans les différentes études modélisant l'activité. Les modélisations présentées diffèrent selon la nature de l'activité (statique ou dynamique) mais également selon le contexte d'étude, ce qui représente une limite et une difficulté majeure pour la réutilisation de ces modèles.

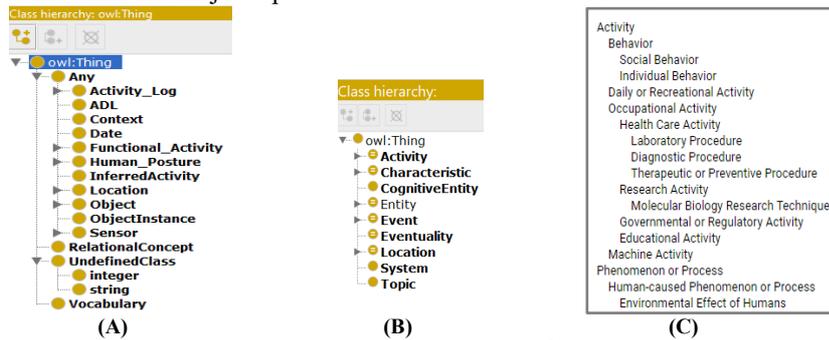

**Figure 1 :** *Capture de l'ontologie « Home Activity[1] » (A) (Wongpatikaseree et al., 2013), de l'ontologie DBpedia[2] (B), et du méta-thésaurus UMLS[3] (C) faisant notion au concept « Activity ».*

|   | Activity | Agent | Location | Effects | Enviromental Entity | Condition | Goal | DateTime | Duration | Action | Person | Context | Object | Sensor | Entity | Requirement | Outcome | Caracteristic | Cognitive Entity | Event | Eventuality | System | Topic |
|---|---|---|---|---|---|---|---|---|---|---|---|---|---|---|---|---|---|---|---|---|---|---|---|
| 1 | x | x | x | x | x | x | x | x | x |   |   |   |   |   |   |   |   |   |   |   |   |   |   |
| 2 | x |   |   |   |   |   |   | x |   | x | x |   |   |   |   |   |   |   |   |   |   |   |   |
| 3 | x |   | x |   |   |   |   | x |   |   |   | x | x | x |   |   |   |   |   |   |   |   |   |
| 4 | x | x |   |   |   |   |   |   |   |   |   |   |   |   | x |   |   |   |   |   |   |   |   |
| 5 | x |   | x |   |   |   |   |   |   |   |   |   |   |   | x |   |   | x | x | x | x | x | x |
| 6 | x | x | x | x |   |   |   |   |   |   |   |   |   |   |   | x | x |   |   |   |   |   |   |
| 7 | x | x | x |   | x |   |   | x | x |   |   |   |   |   |   |   |   |   |   |   |   |   |   |

**Tableau 1 :** *Synthèse des principaux termes identifiés (niveau 1 de la ressource, énumérés par ordre de citation) reliés au concept « Activity ». 1 : Chen et al., 2008 ; 2 : Riboni and Bettini, 2011 ; 3 : Wongpatikaseree et al., 2013 ; 4 : PROV-O, 2013 ; 5 : DBpedia, 2014 ; 6 : Abdalla et al., 2016 ; 7 : Hoda et al., 2018.*

Nous nous sommes également intéressés aux relations explicitées entre les différents termes identifiés. Le Tableau 2 synthétise les principaux termes en relation avec le concept « Person » trouvés dans les différentes études présentées

---
[1] https://lov.linkeddata.es/dataset/lov/vocabs/ha
[2] http://dbpedia.org/ontology/Activity
[3] https://www.nlm.nih.gov/research/umls/

*supra*. Nous pouvons observer que contrairement aux concepts, aucune relation n'est réutilisée et que (Wongpatikaseree *et al.*, 2013) ainsi que la ressource DBpedia[2] ne présentent dans leur modèle aucune relation. Nous pouvons observer dans le Tableau 1 que le concept « Person » n'est pas toujours inclus dans les différentes représentations étudiées et que lorsque ce concept est considéré, sa représentation dépend du domaine étudié. Cependant, la suggestion d'activité ne peut être aboutie et personnalisée que si elle prend en considération le profil de la personne. Le Tableau 3 résume les principaux termes de premier niveau reliés au concept « Activity ». Parmi les études citées, plusieurs termes sont utilisés pour conceptualiser la personne : Person, Profil, Agent, User. Nous observons également que les termes utilisés dépendent fortement du contexte étudié, d'où la nécessité de construire un profil permettant organiser les connaissances afin d'adapter la suggestion d'activités aux PA.

| | superActivity | subActivity | typeOf | performed | atLocation | require | hasEntities | lastFor | achieve | resultedIn | PerformedDuring | impliesAction | PerformsActivity | hasArtifact | isInside | includeCommunica-tionRoute | actedOnBehalfOf | wasAssociatedWith | wasAttributedTo | used | wasGeneratedBy | startedAtime | endedAtTime | wasDerivedFrom | wasInformedBy |
|---|---|---|---|---|---|---|---|---|---|---|---|---|---|---|---|---|---|---|---|---|---|---|---|---|---|
| 1 | x | x | x | x | x | x | x | x | x | x | | | | | | | | | | | | | | | |
| 2 | | | | | | | | | | | x | x | x | x | x | x | | | | | | | | | |
| 4 | | | | | | | | | | | | | | | | | x | x | x | x | x | x | x | x | x |

| | takePlaceAt | hasEnd | hasDuration | Produces | hasPart | has Participant | has Requirement | hasTemporal Description | hasEnvirement Entity | hasLocation | Perform |
|---|---|---|---|---|---|---|---|---|---|---|---|
| 6 | x | x | x | x | x | x | x | | | | |
| 7 | | | | | | | | x | x | x | x |

**Tableau 2** : *Synthèse montrant les relations utilisées reliés les différents termes définis par les auteurs cités dans le Tableau 1.1 : Chen et al., 2008 ; 2 : Riboni and Bettini, 2011 ; 4 : PROV-O, 2013 ; 6 : Abdalla et al., 2016 ; 7 : Hoda et al., 2018.*

La seule ontologie de l'activité accessible depuis BioPortal est l'ontologie SMASH (*Semantic Mining of Activity, Social, and Health data*)[4]. Cette ontologie est destinée aux réseaux sociaux de santé et est fondée sur l'étude YesiWell (Phan *et al.*, 2015). Elle est composée de trois autres ontologies et définit des concepts associés à une perte de poids durable, en particulier ceux liés à une intervention continue avec de fréquents contacts sociaux. Le groupe de travail du W3C fournit quant à lui une ontologie PROV-O (PROVenance, 2013) minimale qui est composée de trois concepts centraux : « *Activity* », « *Entity* » et « *Agent* » et constitue une ressource intéressante. Nous constatons que parmi toutes les ontologies citées, aucune ne rassemble toutes les activités pouvant être exercées. Depuis la fin des années 80, un compendium des activités physiques (*Compendium of Physical Activities*, CPA)[5] pour adultes a été développé pour être utilisé dans des études épidémiologiques afin de normaliser l'attribution des intensités mesurées par l'échelle métabolique (*Metabolic Equivalent of Task*, MET) dans les questionnaires sur l'activité physique. Sa dernière mise à jour date de 2011 (Ainsworth *et al.*, 2011).

---
[4] https://bioportal.bioontology.org/ontologies/SMASH

| | Person | Profile | Preferences | BasicPersonal Information | User | Role | Physiological Caracteristics | UsageStatistics | BasicHealth Information | Medical Information | Culture | Religion | Risk | Health Condition | LifeStyle | Goal | FitnessLevel | Plan | Interests | Identity | Physiological State | Location | Device |
|---|---|---|---|---|---|---|---|---|---|---|---|---|---|---|---|---|---|---|---|---|---|---|---|
| 1 | | x | x | x | | x | | | | | | | | | | | | | | x | | | |
| 2 | | | x | | x | | x | x | x | x | x | x | x | x | | | | | | | | | |
| 3 | x | | | | | | | | | | | | | | x | | | | | | | | |
| 4 | x | x | x | | | x | | | | | | | | | | | | | | | | | |
| 5 | x | | x | x | | | x | | | | | | | | | x | x | x | x | | x | x | x |

**Tableau 3 :** *Synthèse des principaux termes identifiés dans les différentes études explorées associés au concept « Person ». 1 : Primo et al., 2012 ; 2 : Al-nazer et al., 2014 ; 3 : Yu et al., 2015 ; 4 : Ni et al., 2016 ; 5 : Hoda et al., 2018.*

## 3. L'ontologie OAPA

L'ontologie de l'Activité pour les Personnes Agées (OAPA) (*Ontology of Activity For the Elderly* (OAFE)) a été conçue pour modéliser le domaine de l'activité dans le but d'adapter la suggestion d'activité en fonction de l'état de santé, du niveau d'activité, du niveau d'autonomie, de l'environnement et des préférences de la PA. L'ontologie décrit les caractéristiques propres à l'activité et le contexte dans lequel elle se déroule. Le projet NeON[6] (Suárez-Figueroa *et al.*, 2012) constitue le cadre méthodologique dans lequel nous nous situons pour construire l'ontologie. Les scénarios 1 (spécification, planification, conceptualisation, formalisation, implémentation) et 4 (réutilisation) ont été appliqués. Le périmètre de l'ontologie est établi en adoptant la méthodologie de (Uschold and Gruninger,1996).

### 3.1. *Scénarios et questions de compétences*

Afin de simplifier la gestion de l'ontologie en cours de développement, de contrôler l'impact de son évolution et d'assurer la consistance de l'ontologie, nous avons établi un certain nombre de scénarios. Ces scénarios permettent de définir le périmètre de l'ontologie. Ce dernier couvre le domaine des activités favorisant le déplacement, limitant la sédentarité et améliorant l'état de santé et le bien-être des PA non dépendantes, sédentaires et/ou inactives vivant à domicile.

Nous présentons dans le Tableau 4 une liste non exhaustive des scénarios issus principalement de cas d'utilisation identifiés à partir des discussions de forums.

---

[5] https://sites.google.com/site/compendiumofphysicalactivities/home

Les questions de compétences (QC) consistent en un ensemble de questions auxquelles une ontologie devrait pouvoir répondre selon un scénario d'usage (Grüninger and Fox, 1995). Un large éventail de QC, leur utilité dans la création d'ontologies et leur intégration possible dans des outils de création ont été étudiés (Ren *et al.*, 2014 ; Hofer *et al.*, 2015 ; Dennis *et al.*, 2017). Traditionnellement, les QC sont utilisées lors du développement d'ontologies pour rassembler les besoins fonctionnels des utilisateurs dans des cas d'utilisation spécifiques garantissant que toutes les informations pertinentes sont codées (Jacobson *et al.*, 1992). Les QC constituent ainsi un moyen pour déterminer les spécifications de l'ontologie à concevoir et d'évaluer un modèle de connaissances. Par exemple, une réponse à une QC peut être obtenue simplement en parcourant la hiérarchie des concepts, ou en visualisant les relations entre les entités de l'ontologie, ou en déterminant l'héritage, ou par d'autres mécanismes (interrogation d'ontologie, etc.).

| N° | Scénario d'usage | L'ontologie doit permettre de représenter |
|---|---|---|
| 1 | La PA s'interroge sur les activités qu'elle pourrait pratiquer en fonction de son état de santé et de ses pathologies. | les activités en lien avec un état de santé et une pathologie donnée. |
| 2 | La PA s'interroge sur les activités physiques qu'elle peut pratiquer en fonction de son niveau d'autonomie. | les activités physiques en décrivant le niveau d'autonomie nécessaire pour les pratiquer. |
| 3 | La PA s'interroge sur les activités physiques qu'elle peut pratiquer en fonction de ses habitudes de vie. | les activités physiques en décrivant les liens avec le mode de vie de la PA. |
| 4 | La PA s'interroge sur les activités qu'elle peut pratiquer en fonction des objectifs qu'elle se fixe (améliorer son endurance, etc). | les activités physiques en caractérisant les bénéfices qu'elle procure (marcheNordique aPourGainPhysique Endurance). |
| 5 | La PA s'interroge sur la localisation où se déroule une activité physique et le matériel qu'elle doit se procurer pour l'exercer. | les localisations et les matériels associés à chaque activité physique. |

**Tableau 4 :** *Exemples de scénarios d'usage.*

Nous dressons ci-dessous une liste non exhaustive de QC issues de forums, de discussions sur des sites tels que l'EHPAD. L'ontologie doit être en mesure de répondre à des questions comme ceux énoncés dans le Tableau 5. Chaque QC est reliée au minimum à un scénario d'usage (S) (cf. Tableau 4).

| N° | QC | S |
|---|---|---|
| 1 | J'ai 85 ans et j'ai de l'hypertension artérielle depuis quelques années, j'aimerai savoir si la marche nordique est une activité qui peut avoir des bienfaits sur ma maladie ? | 1, 3 |
| 2 | Quels sont les activités physiques qui pourraient participer à maintenir mon indépendance fonctionnelle sachant que je suis à la retraite et que mon activité est, jour après jour, de plus en plus réduite ? | 2, 4 |
| 3 | J'ai 62 ans, il m'a été diagnostiqué de la polyarthrite rhumatoïde, quels exercices dois-je pratiquer pour maintenir un maximum de mobilité et préserver mes articulations ? | 3 |
| 4 | J'ai subi une intervention pour une hernie discale il y a un mois, y a t-il des risques de faire du ski de fond sans risque de récidives ? | 1, 3 |
| 5 | J'ai 65 ans et j'aimerai savoir s'il y a des bienfaits associés à la gymnastique douce si je l'exerce uniquement une fois par semaine et quels sont les matériels que je dois avoir à ma disposition ? | 3, 5 |

**Tableau 5 :** *Exemples de questions de compétences (QC) associées aux scénarios (S).*

### 3.2. Concept Activité

L'étape d'acquisition des connaissances a été réalisée à partir de sources officielles (OMS, PNNS) et de sites spécialisés tels que les établissements d'hébergement pour personnes âgées dépendantes (EHPAD) fournissant des lignes directives de recommandations d'activités physiques pour les PA ainsi que les bienfaits associés à ces activités. L'acquisition de connaissances en plus de l'élaboration des scénarios et des QC ont permis de construire un modèle conceptuel construit à partir d'une carte conceptuelle. Les activités ont été définies à partir du vocabulaire CPA permettant d'associer l'intensité (mesurée à l'aide de l'échelle MET), le rythme (RythmeFaible, RythmeModere, etc.) et l'effort fourni (EffortLeger, EffortIntense, etc.) au moment de l'activité. Pour cela, l'intensité, le rythme et l'effort ont nécessité une grande attention. Par exemple, la mesure de l'effort est loin d'être une information évidente à définir et à modéliser. En effet, l'effort présente plusieurs dimensions, une dimension physiologique (effort physique), une dimension psychologique (effort mental) et une dimension plus large incluant l'effort perçu. Ceci implique la définition des déterminants de l'effort (débit d'oxygène (VO2), fréquence cardiaque (Fc), etc.) mesurés à partir de l'échelle RPE (*Rating of Perceived Exertion*) de Borg pour la détermination de l'intensité de l'activité (Borg, 1998). La présence d'une linéarité entre l'intensité de l'exercice, les contraintes physiologiques (VO2, Fc) et l'effort perçu nous permet de compléter la définition de l'intensité d'une activité.

### 3.3. *Réutilisation d'ontologies*

Au cours du scénario de réutilisation, l'ontologie SMASH ainsi que le vocabulaire PROV-O ont été choisis pour être réutilisés lors de la conception de OAPA. A partir des modélisations et ontologies sélectionnées, le vocabulaire des ontologies candidates dans lequel les primitives ontologiques ont été définies (concepts, relations et axiomes) a été réutilisé. Les autres ontologies citées n'ont pas été réutilisées soit en raison de leur contexte trop éloigné, soit en raison du manque de granularité du modèle, soit en raison de l'absence de la ressource ontologique. L'ontologie *Human Disease Ontology*[7] a également été réutilisée pour définir les principales classes des pathologies et des symptômes utilisés afin de construire le profil de la PA et personnaliser la suggestion d'activité.

### 3.4. Concept Personne

La personnalisation de la suggestion nécessite la prise en compte du profil de la PA. Pour cela, un large éventail d'information devra être recueilli auprès d'elle. Ces informations sont fixes ou variables. Les données fixes d'une personne (Nom, Taille, etc.) sont acquises une seule fois et ne varient pas en fonction du temps. Elles sont modélisées sous le concept « Identite ». Certaines propriétés peuvent soit varier au cours du temps (Poids, etc.) soit être contextuelles (FrequenceCardiaque, PressionArterielle, etc.). Ainsi, toutes les variables impliquées dans la définition de l'état de santé (CaracteristiquePhysiologique, Symptome, Pathologie), du profil de l'activité et de l'environnement social (NiveauActivite, CaracteristiqueSocial, etc.), des objectifs que la personne souhaite ou devrait atteindre (Objectif) ont été décrites (Figure 3 (A)). Cette modélisation a été enrichie en associant les freins liés à l'état de santé, au profil d'activité ou encore au profil social (Alfaifi *et al.*, 2017) de la PA.

---
[7] https://bioportal.bioontology.org/ontologies/DOID

### 3.5. Présentation de l'ontologie OAPA

La modélisation des différentes cartes conceptuelles a servi de base à la construction de l'ontologie formalisée en OWL2. Afin de mieux spécifier la signification de chaque classe, nous nous sommes référés à d'autres ontologies standardisées telles que la nomenclature systématique des termes cliniques de la médecine (SNOMED-CT[8]) en rapport avec l'activité et des entités permettant de définir un état de santé pour la personne. L'éditeur Protégé est utilisé pour construire l'ontologie et réutiliser les vocabulaires d'intérêts.

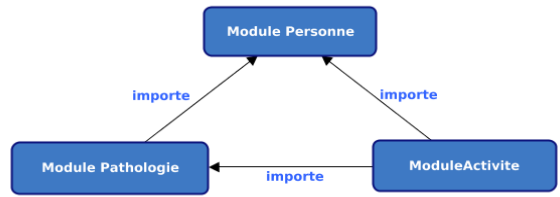

**Figure 2 :** *Schéma montrant la structuration des modules importés.*

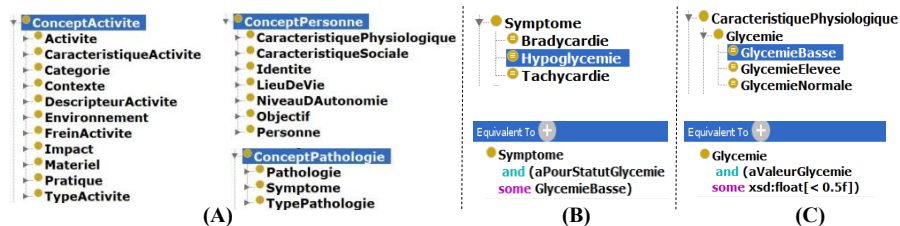

**Figure 3 :** *Aperçu de la hiérarchie des classes de OAPA (A) et d'un exemple sur la définition des classes équivalentes concernant l'hypoglycémie (B) et(C).*

OAPA a été conçue de façon modulaire suivant une approche par composition. Chaque module correspondant à un domaine a été conçu pour être auto-suffisant. Le module Personne est un module auquel les autres modules font référence (Figure 2). Dans Protégé, la relation d'import a été utilisée pour caractériser les activités de la PA. OAPA est actuellement composée de 3 modules (ConceptActivite, ConceptPersonne, ConceptActivite), 405 classes, de 61 « *object property* » et de 42 « *data property* ». Les concepts du modèle conceptuel (Figure 3 (A)) sont représentés par des classes et les relations par des propriétés (« *object property* », « *data property* »). Chaque entité est annotée par un label, un altlabel, et une définition. L'exemple de la Figure 3 (B) et (C) montre la définition de la classe équivalente « Hypoglycemie » et de « GlycemieBasse ». Cette restriction et assertion entre les classes et les individus nous permettra d'attribuer par inférence un symptôme (Hypoglycemie) et un statut glycémique (GlycemieBasse) selon la valeur glycémique (aValeurGlycemie) de la PA.

---

[8]http://bioportal.bioontology.org/ontologies/SNOMEDCT/

## 4. Evaluation

Pour évaluer OAPA et s'assurer de sa consistance et de sa cohérence, une série de requêtes DL Query ont été faites pour répondre aux QC notamment celles citées dans le Tableau 5. La Figure 4 présente un exemple de requêtes permettant de répondre à la question suivante : Quelles sont les activités de faible intensité ayant un impact sur l'endurance et s'exerçant à l'extérieur ? Parmi les activités sélectionnées pour cet exemple, nous pouvons constater qu'à chaque restriction, le nombre de choix parmi les activités diminue. Ces résultats ont pu être obtenus grâce aux classes définies utilisées au moment de ces requêtes. Ainsi, les classes définies, comme ActiviteEndurance, permettent d'utiliser le raisonneur (Fact++) pour caractériser les activités selon les contraintes définies.

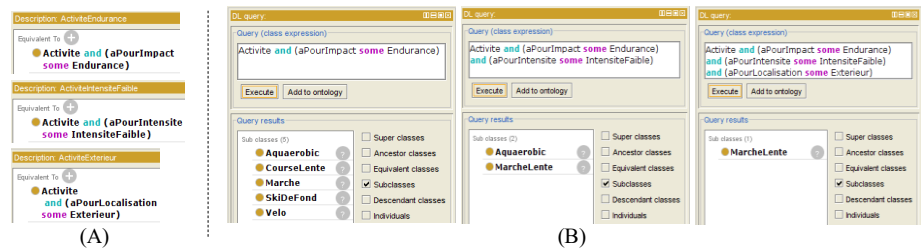

**Figure 4 :** *Aperçu de quelques exemples de classes définies (A) et de requêtes (DL Query) (B) pour obtenir les activités correspondantes à la requête exécutée.*

## 5. Conclusion

Dans ce papier, nous avons présenté un travail préliminaire pour la modélisation et le développement d'une Ontologie de l'Activité pour les Personnes Agées (OAPA) pour la suggestion d'activités personnalisées, ciblées et incitatives pour les PA sédentaires, dépendants ou en perte d'autonomie. La rigueur dans la représentation des activités et de la personne est une base nécessaire pour une interopérabilité, un échange et une réutilisation significative des données issues de capteurs. En perspective de ce travail, nous souhaitons inclure dans cette modélisation les objets connectés et les services pour être en mesure de concevoir un système de suggestion d'activités aux PA à partir des données collectées la concernant. L'utilisation d'objets connectés à condition qu'ils soient acceptés et bien utilisés facilitera la collecte des données de santé et de l'activité. L'approche ontologique adoptée nous permettra d'exploiter les relations entre les concepts modélisés pour raisonner sur les activités représentées dans OAPA. En rendant OAPA accessible pour tous et en la traduisant dans d'autres langues, notre travail participera à l'évolution de cette thématique en pleine expansion et prometteuse pour les années à venir.

## 5. Bibliographie

Abdalla A., Hu Y., Carral D., Li N., Janowicz K., « An ontology design pattern for activity reasoning », *Proceedings of the 5th International Conference on Ontology and Semantic Web Patterns*, Aachen, Germany, vol. 1302,2014, p. 78-81.

Ainsworth BE., Haskell WL., Herrmann SD., Meckes N., Bassett DR., Tudor-Locke C., « 2011 Compendium of Physical Activities: A second update of codes and met values», *Medicine & Science in Sports & Exercise*, vol. 43, n° 8, août 2011, p. 1575-81.

Alfaifi Y., Grasso F., Tamma V., « Towards an ontology to identify barriers to physical activity for type 2 diabetes », *Proceedings of the 2017 International Conference on Digital Health, New York (DH '17)*, USA, 2017, p. 16–20.

Al-Nazer A., Helmy T., Al-Mulhem M., « User's profile ontology-based semantic framework for personalized food and nutrition recommendation », *Procedia Computer Science*, vol. 32, 2014, p. 101-8.

Borg G., *Borg's Perceived exertion and pain scales*, Champaign, Human Kinetics, 1998.

Chen L., Nugent C., « Ontology-based activity recognition in intelligent pervasive environments »,*International Journal of Web Information Systems*, vol. 5, n° 4, 20 novembre 2009, p. 410-30.

Dennis M., van Deemter K., Dell'Aglio D., Pan JZ., « Computing Authoring tests from competency questions: experimental validation», *International Semantic Web Conference (ISWC 2017)*, 2017, p. 243-59.

Escalon H., Beck F., « Perceptions, connaissances et comportements en matière d'alimentation. Les spécificités des seniors »,*Gérontologie et société*, vol. 33, n° 3,20 octobre 2010, p. 13-29.

Filos D., Triantafyllidis A., Chouvarda I., Buys R., Cornelissen V., Budts W.,Woods C., Moran K., Maglaveras N., « PATHway: Decision Support in exercise programmes for cardiac rehabilitation », *Studies in Health Technology and Informatics*, vol. 224,2016, p. 40-5.

Gayathri KS., Easwarakumar KS., Elias S., « Probabilistic ontology based activity recognition in smart homes using Markov Logic Network», *Knowledge-Based Systems*, vol. 121, avril 2017, p. 173-84.

Grüninger M., Fox MS., « Methodology for the design and evaluation of ontologies », *Department of Industrial Engineering*, 1995.

Hoda M., Montaghami V., Al Osman H., El Saddik A., « ECOPPA: Extensible Context Ontology for Persuasive Physical-Activity Applications», *Proceedings of the International Conference on Information Technology & Systems (ICITS 2018)*, 2018, p. 309-18.

Hofer P., Neururer S., Hauffe H., Insam T., Zeilner A., Gobel G., « Semi-automated evaluation of biomedical ontologies for the biobanking domain based on competency questions», *Studies in Health Technology and Informatics*, 2015, p. 65–72.

Nassabi MH., op den Akker H., Vollenbroek-Hutten M., « An ontology-based recommender system to promote physical activity for pre-frail elderly», *Mensch & Computer*, janvier 2014.

Ni Q., Pau de la Cruz I., García Hernando AB., « A foundational ontology-based model for human activity representation in smart homes», *Journal of Ambient Intelligence and Smart Environments*, vol. 8, n° 1, 7 janvier 2016, p. 47-61.

Phan N., Dou D., Wang H., Kil D., Piniewski B., « Ontology-based deep learning for human behavior prediction with explanations in health social networks», *Information Sciences*, vol. 384, avril 2017, p. 298-313.

Pramono D., Setiawan NY., Sarno R., Sidiq M., « Physical activity recommendation for diabetic patients based on ontology», *7th International Conference on Information & Communication Technology and Systems*, 2013, p. 27– 32.

Primo T., Silva JLT., Ribeiro AM., Vicari RM., Boff E., « Towards ontological profiles in communities of practice», *IEEE Multidisciplinary Engineering Education Magazine*, vol. 7, 2012, p. 13–22.

Ren Y., Parvizi A., Mellish C., Pan JZ., Deemter K van., Stevens R., « Towards competency question-driven ontology authoring», *The Semantic Web*, 2014, p. 752-67.

Riboni D., Bettini C., « OWL 2 modeling and reasoning with complex human activities », *Pervasive and Mobile Computing*, vol. 7, n° 3, juin 2011, p. 379-95.

Suárez-Figueroa MC., Gómez-Pérez A., Fernández-López M., « The NeOn Methodology for Ontology Engineering », *Ontology Engineering in a Networked World*, Berlin, 2012, p. 9-34.

Uschold M., Gruninger M., « Ontologies : principles, methods and applications », *The Knowledge Engineering Review*, vol. 11, n° 2, juin 1996, p. 93.

Wieringa W., op den Akker H., Jones VM., op den Akker R., Hermens HJ., « Ontology-based generation of dynamic feedback on physical activity », *Artificial Intelligence in Medicine*, Berlin, Heidelberg, 2011, p. 55-9.

Wongpatikaseree K., Ikeda M., Buranarach M., Supnithi T., Lim AO., Tan Y., « Location-based concept in activity log ontology for activity recognition in smart home domain», *Joint International Semantic Technology Conference (JIST) : Semantic Technology*, Berlin, Heidelberg, 2013, p. 326-31.

Yu HQ., Zhao X., Deng Z., Dong F.. «Ontology Driven Personal Health Knowledge Discovery », *Knowledge Management in Organizations*, 2015, p. 649-63.

Zolfaghari S., Zall R., Keyvanpour MR.., « SOnAr: Smart Ontology activity recognition framework to fulfill semantic web in smart homes», *Second International Conference on Web Research (ICWR)*, 2016, p. 139-44.